\documentclass[10pt,twocolumn,letterpaper]{article}

\usepackage{iccv}
\usepackage{times}
\usepackage{epsfig}
\usepackage{graphicx}
\usepackage{amsmath}
\usepackage{amssymb}
\usepackage{booktabs}
\usepackage{multirow, makecell}
\usepackage[table, dvipsnames]{xcolor}
\usepackage{pgfplots}
\usepackage[nocompress]{cite}
\usepgfplotslibrary{fillbetween}

\usepackage[pagebackref=true,breaklinks=true,letterpaper=true,colorlinks,bookmarks=false]{hyperref}

\iccvfinalcopy

\begin{document}

%%%%%%%%% TITLE
\title{GACE: Geometry Aware Confidence Enhancement for \\ Black-box 3D Object Detectors on LiDAR-Data}
\author{
    {David Schinagl\textsuperscript{1,2} Georg Krispel\textsuperscript{1} Christian Fruhwirth-Reisinger\textsuperscript{1,2} Horst Possegger\textsuperscript{1} Horst Bischof\textsuperscript{1,2}}\\
        {\tt\small\{david.schinagl,georg.krispel,reisinger,possegger,bischof\}@icg.tugraz.at} \\
        \textsuperscript{1} Graz University of Technology \, \textsuperscript{2} Christian Doppler Laboratory for Embedded Machine Learning \\
}

\maketitle
\thispagestyle{empty}

%%%%%%%%% ABSTRACT
\begin{abstract}
Widely-used LiDAR-based 3D object detectors often neglect fundamental geometric information readily available from the object proposals in their confidence estimation.
This is mostly due to architectural design choices, which were often adopted from the 2D image domain, where geometric context is rarely available.
In 3D, however, considering the object properties and its surroundings in a holistic way is important to distinguish between true and false positive detections, \eg occluded pedestrians in a group.
To address this, we present GACE, an intuitive and highly efficient method to improve the confidence estimation of a given black-box 3D object detector.
We aggregate geometric cues of detections and their spatial relationships, which enables us to properly assess their plausibility and consequently, improve the confidence estimation.
This leads to consistent performance gains over a variety of state-of-the-art detectors.
Across all evaluated detectors, GACE proves to be especially beneficial for the vulnerable road user classes, \ie~pedestrians and cyclists.
\end{abstract}

%%%%%%%%% BODY TEXT
\section{Introduction}
\label{sec:intro}

\begin{figure}[t]
	\centering
	\includegraphics[width=0.98\linewidth]{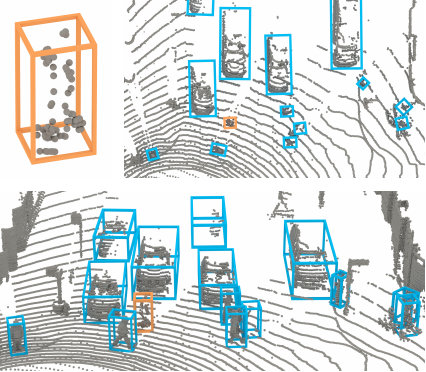}
	\caption{
		Context matters: a baseline detector struggles at detecting true positive objects confidently if the sampling pattern is atypical, \eg~the occluded pedestrian in the orange bounding box (close-up top-left).
		GACE exploits the geometric properties of the detection and its surrounding objects to significantly increase the score for this detection, which is intuitively correct for a human observer considering this scene.}
	\label{fig:intro}
\end{figure}

Three-dimensional perception of surrounding objects is a critical component for autonomous vehicles and robots.
Many modern perception systems use point cloud data from LiDAR (Light Detection and Ranging) sensors for this task, since they can provide accurate 3D information even over long distances.
The popularity of these sensors can be seen in the increased research interest in LiDAR-based 3D object detection approaches, \eg~\cite{3dod:iou:hu:afdetv2, 3dod:iou:zheng:CIA-SSD, 3dod:pb:yang:3DSSD, 3dod:gb:zhou:centerformer,3dod:hy:shi:pvrcnnpp, 3dod:gb:yan:SECOND} and the large number of recently published autonomous driving datasets that include LiDAR data, \eg~\cite{3dod:data:sun:waymo, 3dod:data:geiger:kitti, 3dod:data:caesar:nuscenes, 3dod:data:mao:once}.

However, the characteristics of LiDAR data impose significant challenges for object detection.
Unlike pixels in an image, which are aligned in a regular grid, point clouds represent 3D data as a collection of individual points in space, each with its own set of coordinates.
In addition to the unstructured nature of the data, the highly variable point density poses a major challenge.
Due to the angular offset of the LiDAR beams, the density is highly dependent on the distance to the object and can be altered by occlusions in the foreground.
This often requires detecting objects based on very few data points, which is especially true for classes with smaller spatial dimensions, such as pedestrians and cyclists.
For example, in the Waymo Open Dataset~\cite{3dod:data:sun:waymo}, one of the most widely used and challenging datasets to date, about 30 percent of all annotated pedestrians consist of less than merely 20 points.
Detecting these sparsely sensed objects naturally leads to a large number of false positive detections at test time.
For this reason, determining a meaningful confidence value for the detections is critical to find a trade-off between precision and recall that adequately distinguishes true positives from false positives.
The potential that could be exploited by improving the confidence score is considerable:
Suppose we have an oracle that could correctly classify the detections of a SECOND~\cite{3dod:gb:yan:SECOND} model on the Waymo dataset into true and false positives.
This would increase the LEVEL\_1 average precision for vehicles by $+3.96$AP and, more importantly, for pedestrians and cyclists by as much as $+10.71$AP and $+13.74$AP, respectively.

Existing 3D object recognition pipelines, including confidence estimation approaches, were largely inspired by 2D image-based object recognition models and then gradually adapted to point cloud processing.
However, the conventional \textit{backbone - neck - head} architecture of the 2D detection model was largely retained, \eg~in~\cite{3dod:hy:he:structureaware, 3dod:iou:hu:afdetv2, 3dod:gb:lang:PointPillars, 3dod:gb:yan:SECOND,3dod:iou:zheng:CIA-SSD, 3dod:gb:shi:PillarNet}.
After extracting features (point-based, voxel-based, or region-based) over multiple levels in the backbone, they are fused in the neck module and then passed to the detection head, where bounding box regression and confidence estimation are performed on a dense feature representation.
Typically, separate branches are used within the head for bounding box regression and confidence estimation, each consisting of one or more fully connected layers on top of the common feature representation.

\begin{figure}
	\centering
	\includegraphics[width=0.975\linewidth]{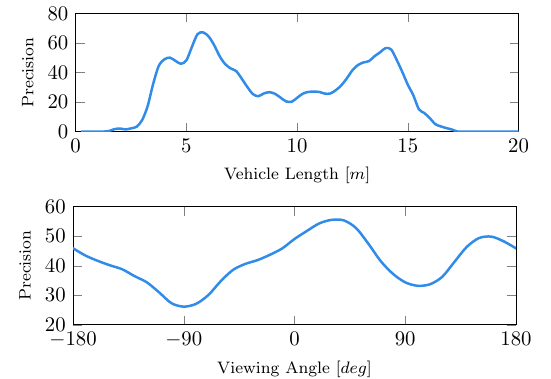}
	\caption{ Precision of a SECOND~\cite{3dod:gb:yan:SECOND} model for the Waymo~\cite{3dod:data:sun:waymo} vehicle class as a function of the object length (top) and of the viewing angle (bottom), indicating from which side the vehicle is seen by the LiDAR, where 0 degrees corresponds to the rear view. These examples illustrate the strong dependence of the precision on simple geometric object properties.}
	\label{fig:geometry}
\end{figure}

Unlike image-based object detection, there are several highly relevant geometric properties inherent to objects in 3D point clouds that have been largely unexploited in assessing the confidence of a detection.
In the image case, it is usually not possible to easily derive geometric properties for the objects, such as height or orientation, unless a static and fully calibrated camera and a known or constant scene are given.
In contrast, in 3D object detection, many geometric features are directly available in the object properties and associated 3D data points to better assess the presence of a real object.
On the one hand, these are \textbf{instance-specific properties}, such as the dimension of the object, the heading direction, the position, or the point distribution within the bounding box.
For example, as shown in Figure~\ref{fig:geometry}, the precision of a SECOND~\cite{3dod:gb:yan:SECOND} model for detecting a vehicle is highly dependent on the size of the object and from which side it is detected by the LiDAR.
It can be seen that vehicle categories between passenger cars and heavy duty vehicles (\ie vehicles with a length of 6 to 13m) are harder to detect, potentially because they are underrepresented in the dataset, and that vehicles are easier to detect from behind (\ie viewing angle between $\pm 45$ degrees), presumably because of the highly reflective license plates, as shown in~\cite{3dod:xai:schinagl:occam}.
On the other hand, \textbf{contextual properties}, \ie geometric relationships to neighboring objects, can contribute to a more reliable estimation of the confidence. For example, as shown in Figure~\ref{fig:intro}, a pedestrian that appears atypical due to occlusions can be assessed more reliably by additionally taking neighboring vehicles and pedestrians into consideration.\\

Nevertheless, these simple but highly informative metric properties are neglected when estimating the confidence score in current detector architectures for the following reasons:
First, in grid-based models, information such as the exact point distribution within the bounding box or the number of points is already partially discarded by the discretization (voxelization) in the preprocessing phase.
Second, the bounding box properties (object dimensions \& rotation) are determined in the parallel and separate box regression head and are therefore not accessible to the confidence estimation head.
Finally, confidence estimation is usually performed using only features within a small area around the object (depending on the receptive field and detector), and no explicit information about neighboring objects and their geometric properties or confidence values is used, preventing a holistic estimation.

Inspired by these observations, we present \emph{GACE}, an intuitive and highly effective method to improve the confidence estimation of any black-box detector using geometric information.
Given a set of detections from the base detector, we explicitly use these neglected features to enhance the expressiveness of the confidence values with the help of these additional cues.
Our model-agnostic approach is intentionally applied after the actual detector training process to perform an auxiliary geometric assessment independent of the initial features.
In a detailed evaluation on the Waymo dataset, we show that GACE consistently improves the performance of several state-of-the-art detection pipelines.
Furthermore, we demonstrate the generalizability and transferability of our method by applying it to other datasets and even other detectors.
Without retraining them, we achieve highly compelling performance gains.

\section{Related Work}
\label{sec:rel_work}

\noindent\textbf{LiDAR-based 3D Object Detection: } Depending on how existing methods for 3D object detection in single frame LiDAR data deal with the unstructured nature of point clouds, they can be broadly categorized into point-based, grid-based, and hybrid approaches.

\noindent\textbf{Point-based methods} extract information directly from the individual raw 3D points~\cite{3dod:pb:qi:pointnet, 3dod:pb:qi:pointnet++, 3dod:pb:shi:Point-RCNN, 3dod:pb:yang:3DSSD, 3dod:pb:shi:Point-GNN, 3dod:pb:zhang:Not-All-Points}.
The pioneering works, PointNet~\cite{3dod:pb:qi:pointnet} and PointNet++~\cite{3dod:pb:qi:pointnet++}, used shared multilayer perceptrons in combination with global pooling functions to directly extract features from the irregular point cloud data.
In Point-RCNN~\cite{3dod:pb:shi:Point-RCNN} features extracted in this manner are used to segment foreground points and generate proposals based on them.
The advantages of point-based 3D object detectors are that there is no loss of point information due to discretization and the large receptive field, but at the cost of high computational demands.

Instead of processing the points directly, \textbf{grid-based methods}~\cite{3dod:gb:zhou:VoxelNet, 3dod:gb:yan:SECOND, 3dod:gb:lang:PointPillars, 3dod:gb:shi:PillarNet, 3dod:gb:yang:pixor, 3dod:gb:denk:voxelrcnn, 3dod:gb:mao:voxeltransformer, 3dod:gb:sun:SWFormer, 3dod:gb:yin:centerbased} discretize the non-uniform 3D points into regular grids that can then be processed with 2D/3D convolutions. 
Voxelnet~\cite{3dod:gb:zhou:VoxelNet}, a pioneering method, divides the point cloud into uniformly spaced 3D voxels, aggregates information from the points within them and generates predictions using 3D convolutions.
To better handle the large number of empty voxels, SECOND~\cite{3dod:gb:yan:SECOND} introduced sparse 3D convolutions. % to increase speed and improve accuracy.
To reduce complexity, PointPillars~\cite{3dod:gb:lang:PointPillars} and PillarNet~\cite{3dod:gb:shi:PillarNet} use a 2D grid on the ground plane to create a column representation, a single pillar-shaped voxel per location, that can be processed using 2D convolutions. 
As an alternative to such anchor-based methods, Centerpoint~\cite{3dod:gb:yin:centerbased} predicts a bird's-eye view heat map and detects the object center using a keypoint detector.
Recently, transformer-based backbones have also been used to enable long-range relationships between voxels~\cite{3dod:gb:mao:voxeltransformer, 3dod:gb:sun:SWFormer, 3dod:gb:fan:SST, 3dod:gb:zhou:centerformer}.
Object relations within a frame and across multiple frames are mapped by Ret3D~\cite{3dod:hy:wu:ret3D} using a graph and a transformer.
The advantage of grid-based methods is that they can process data faster due to the regular format, but are limited by the loss of point information during the initial discretization phase.

In order to obtain both, multi-scale features and fine-grained information, \textbf{hybrid methods} process voxel and point information jointly~\cite{3dod:hy:shi:PartA2, 3dod:gb:ye:hvnet, 3dod:gb:noh:hvpr, 3dod:gb:yang:STD, 3dod:gb:sheng:CT3D, 3dod:hy:shi:pvrcnn, 3dod:hy:shi:pvrcnnpp}.
PV-RCNN~\cite{3dod:hy:shi:pvrcnn}, uses a set abstraction module that combines surrounding point and voxel features at keypoints to improve the detections. 
Part-A$^2$~\cite{3dod:hy:shi:PartA2} predicts the position of parts within an object based on point features to improve the accuracy, while LiDAR~R-CNN~\cite{3dod:hy:li:lidar-rcnn} uses features of a PointNet~\cite{3dod:pb:qi:pointnet} model that processes points within and around box proposals.
Pyramid~R-CNN~\cite{3dod:hy:mao:pyramidrcnn} creates point features using a pyramid grid structure to acquire fine-grained and long-range contextual information.\\

\noindent\textbf{Confidence Estimation}: In the usual \textit{backbone - neck - head} detector architecture, after the feature extraction and aggregation, the box regression and confidence estimation are performed.
This is usually done in two separate branches based on a common dense feature representation~\cite{3dod:gb:lang:PointPillars, 3dod:gb:yan:SECOND, 3dod:gb:zhou:VoxelNet, 3dod:pb:yang:3DSSD}.
This has the disadvantage that the accuracy of the localization is hardly included in the confidence score.
Inspired by 2D object detection methods~\cite{2dod:iou:tychsen:nms-iou, 2dod:iou:jiang:local-confidence, 2dod:iou:huang:mask-scoring-rcnn}, IoU guided supervision is frequently used to obtain a better correlation between the classification result and the localization accuracy~\cite{3dod:iou:li:gs3d, 3dod:gb:yang:STD, 3dod:iou:zheng:CIA-SSD, 3dod:iou:hu:afdetv2, 3dod:hy:shi:pvrcnn, 3dod:hy:shi:PartA2, 3dod:gb:shi:PillarNet, 3dod:conc:hu:point-density-aware}.
Thereby, the IoU between the predicted box and the ground truth is learned in a third branch during training and then incorporated into the final confidence score at test time.
Hu~\etal~\cite{3dod:conc:hu:point-density-aware} leverage the inherent relationship between object distance and point density to better assess a detection.
Inspired by the 2D approach~\cite{2dod:iou:dai:r-fcn}, a spatial transformation on the feature maps is done by He~\etal~\cite{3dod:hy:he:structureaware} to better align the confidence prediction and bounding box regression.
Related to confidence estimation are also calibration methods~\cite{other:guo:calibration, other:kuppers:od-calibration} where the score should represent a true probability, ~\ie~how likely a detection is.
Detection pipelines, however, aim at the best separation of true and false positives as optimization goal for the confidence prediction.\\

In our confidence estimation method we also pursue this optimization objective, but in contrast to existing approaches, we present a method for refining the confidence values for a given set of detections by exploiting the rich geometric information contained directly in the detections as well as in the underlying 3D points.
While these useful cues for assessing the plausibility of a detection have been largely untapped due to the architecture of common detection pipelines, they allow us to increase the expressiveness of the confidence values.

\section{Geometry-Aware Confidence Enhancement}
\label{sec:method}

\begin{figure*}
	\centering
	\includegraphics[width=1\linewidth]{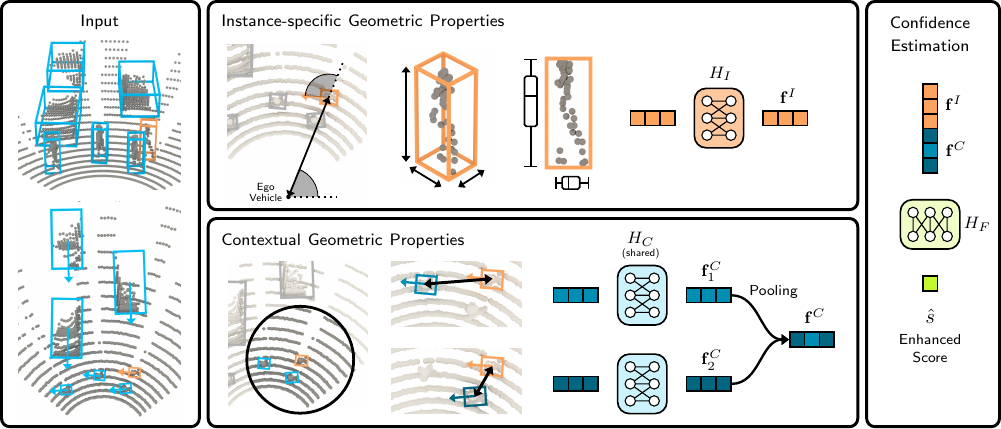}
	\caption{
	Schematic of GACE: To re-evaluate the confidence score of a detection (orange), we aggregate geometric properties of the detection itself and the points it contains into a feature vector (top).
	To capture the context of the detection, geometric relationships to neighboring detections are aggregated using a shared MLP and subsequent pooling function (bottom).
	By merging both features (right), we obtain a new confidence score that takes into account the underlying geometric properties of the detection.
	}
	\label{fig:overview}
	\vspace{-0.4cm}
\end{figure*}

Our goal is to optimize the confidence scores for a given set of detections from a base detector in order to better separate true positives from false positives, thus increasing the overall detection performance.
We use the detector in a pure black box manner, \ie~we do not assume any knowledge about the architecture of the base model, nor access to its internals like parameters, features or gradients.
This black-box optimization, taking only the point cloud and the set of objects detected in it as input, enables universal applicability and easy transferability of our enhancement module to any base 3D object detection pipeline.
The basic idea is to revalidate the detections by exploiting as much as possible the geometric information they inherently contain.\\

In our proposed approach called \textit{GACE} (\textbf{G}eometry \textbf{A}ware \textbf{C}onfidence \textbf{E}nhancement) we exploit two types of geometric information, as shown in Figure~\ref{fig:overview}:
\begin{itemize}
	\item \textbf{Instance-specific Geometric Properties (Section~\ref{sec:method:instance}):} 
	Attributes of the bounding box itself, combined with the point data inside.
	For example, how well is the size of the object or the heading angle supported by the point distribution?
	\item \textbf{Contextual Geometric Properties (Section~\ref{sec:method:contextual}):} Relationships to surrounding objects can provide useful information to better validate an uncertain detection, \eg, a partially occluded vehicle moving in the same lane and same direction as the surrounding vehicles.
\end{itemize}
These useful cues to estimate the plausibility of an object are usually not used for the confidence estimation in common detection pipelines.
The reasons are that information is often already discarded during preprocessing (discretization), essential properties such as the estimated size of the object are not available (separate box regression in a parallel branch), or the objects are only evaluated individually and not more holistically.
Therefore, we explicitly use these easily accessible and rich sources of information as input to our enhancement module.
After merging the instance-specific and contextual features, we determine the new confidence value of each proposed object via an auxiliary task (Section~\ref{sec:method:fusion}).

To generate the training data for our enhancement module, we use the black-box base model in a single inference run on the training set. The resulting set of all detections from the base model represents the training data to learn our improved confidence estimator.
Formally, this can be described as follows.\\

\noindent\textbf{Definitions \& Notations: }
Let $\widetilde{\mathcal{X}}=\{\widetilde{\mathbf{x}}_k\}_{k=1 \ldots K}$ be a LiDAR point cloud, where each of the $K$ unordered points $\widetilde{\mathbf{x}}_k \in \mathbb{R}^5$ consists of the 3D coordinates, intensity/reflectance and the elongation value.
Furthermore, let $\widetilde{\mathcal{Y}} = \{\widetilde{\mathbf{y}}_i\}_{j = 1 \ldots M}$ be the set of corresponding ground truth objects. Each object annotation $\widetilde{\mathbf{y}} = [\widetilde{\mathbf{b}}, \widetilde{y}]$ includes the bounding box parameters $\widetilde{\mathbf{b}} = [\widetilde{c}_x, \widetilde{c}_y, \widetilde{c}_z, \widetilde{d}_x, \widetilde{d}_y, \widetilde{d}_z, \widetilde{\Theta}]$ and the corresponding class label $\widetilde{y}$.
For a given black-box 3D object detector $F$, let ${F(\widetilde{\mathcal{X}})} = {\mathcal{Y} = \{\mathbf{y}}_{i=1 \ldots N}\} $ be the set of proposed detections for this input point cloud, where $\mathbf{y} = [\mathbf{b}, y, s]$.
In addition to the box properties $\mathbf{b}$ and the corresponding class label $y$, the detector predicts a confidence value $s$ that should ideally indicate the prevalence of a true positive example.\\

\noindent\textbf{Confidence Optimization: }
Based on the known ground truth objects, a category label $\{u_i \in \{0,1\}\}_{i = 1 \ldots N}$ can be assigned to each detection, indicating whether it is a true positive or false positive detection.
Moreover, we know the IoU with a possible ground truth bounding box for each object which we define as  $\{v_i\}_{i=1 \ldots N}$.
We aim to improve the confidence estimate of the original detector by focusing exclusively on the binary classification of detections into true positive and false positive examples.
We determine the revised confidence score $\{\widehat{s}_i\} = H(\mathcal{Y},\mathcal{X})$ using our module $H$, where only the set of detections $\mathcal{Y}$ and the points they include $\mathcal{X} \subset \widetilde{\mathcal{X}}$ are used as input.

\subsection{Instance-specific Geometric Properties}
\label{sec:method:instance}

In common 2D and 3D object detection architectures, the bounding box regression and the confidence estimation are performed completely separated in different branches.
This is well suited for 2D where it is desired to detect objects in the image regardless of their scale.
For 3D, however, important cues for plausibility estimation remain largely unused.
Let us consider the available context directly provided by an object proposal:
The position of the possible object in combination with the dimensions indicates, for example, which point density is to be expected.
Furthermore, the direction angle provides an indication of the expected point distribution within the box.
This information can even be further refined by knowing the class of the object.
This directly available geometrical knowledge about an object allows for a more profound estimation of the confidence score.\\

We extract these basic properties and transform them into a compact representation using a multilayer perceptron (MLP) $H_I$.
As input parameters, we first use the object parameters, \ie~the position $(c_x, c_y, c_z)$, the size $(d_x, d_y, d_z)$, and the heading angle $\Theta$ of the bounding box, as well as the initial confidence value $s_i$ of the detection estimated by the base detector.
Additionally, we use the distance $\lVert \mathbf{c} \rVert$ from the LiDAR sensor to the object center, and the angle $\alpha$ between the line of sight to the object and the heading angle of the object:
\begin{equation}
	\alpha = \Theta - \text{atan2}(c_y, c_x).
\end{equation}
This angle describes from which side an object is seen from the LiDAR center, independent of the position of the object relative to the LiDAR, \eg~a vehicle driving directly towards the LiDAR always has the same angle $\alpha$, no matter from which direction the vehicle approaches.
We complement these cues with information about the points $\mathcal{X}_{\mathbf{b}} \subset \mathcal{X}$ inside the bounding box $\mathbf{b}$.
Besides the overall number of points $|\mathcal{X}_{\mathbf{b}}|$, we extract elementary low-level statistics.
Therefore, we scale the box and the associated points to a uniform size and then align them \wrt their center and yaw angle.
In this canonical representation we compute the mean, standard deviation, minimum and maximum of $\mathcal{X}_{\mathbf{b}}$ for all axes, denoted as  $\mathcal{X}_{\mathbf{b}}^{\text{mean}},  \mathcal{X}_{\mathbf{b}}^{\text{std}}, \mathcal{X}_{\mathbf{b}}^{\text{min}}, \mathcal{X}_{\mathbf{b}}^{\text{max}}$.
We then aggregate these attributes into one feature vector representing the instance-specific plausibility per object as
\begin{equation}
	\mathbf{f}^I = H_I\left(\left[ \mathbf{b}, \alpha,  \lVert \mathbf{c} \rVert,  |\mathcal{X}_{\mathbf{b}}|,  \mathcal{X}_{\mathbf{b}}^{\text{mean}},  \mathcal{X}_{\mathbf{b}}^{\text{std}}, \mathcal{X}_{\mathbf{b}}^{\text{min}}, \mathcal{X}_{\mathbf{b}}^{\text{max}}\right]^\top\right),
\end{equation}
where we pass all angles as direction vectors $\left( \cos(\cdot), \sin(\cdot) \right)$ and normalize all metric properties to unit length using the corresponding maximum value range.\\

\subsection{Contextual Geometric Properties}
\label{sec:method:contextual}

Especially in the case of uncertain detections, \eg detections that are far away and therefore consist of only a few 3D points, or detections that are partially occluded and therefore appear atypical, geometric contextual information can be very useful in assessing a confidence score.
Examples include a vehicle that is heavily occluded but in a convoy with other vehicles, or cyclists moving in a group.
However, this information is usually not available for the confidence estimation, since the object proposals are evaluated individually within the receptive field but without explicit knowledge of the objects detected in its vicinity and their properties.
In order to assess the plausibility of a detection $\mathbf{y}$ in a more holistic way, we use the geometric relations to surrounding objects in the scene.
We thereby consider all neighbors that are within a certain radius $r$ around the object to be evaluated.\\

We therefore create a representation per object, which captures the relationships to its neighboring objects.
In order to be independent of the number of neighboring objects, we first create a feature vector for each neighbor, which we then combine into a unified representation using a symmetric pooling function.
As input parameters per neighbor we use the distance to the object to be evaluated $\lVert \mathbf{c} - \mathbf{c}_n \rVert$, the direction vector $\mathbf{c} - \mathbf{c}_n$, the difference between the two heading angles $\Theta - \Theta_n$, and the neighbor's class label $y_n$.
To incorporate the validity of the neighbor, we leverage $\mathbf{f}^I_n$, which represents the instance-specific properties of the neighbor.
This information is encoded via the shared weight MLP $H_C$ to form the feature vector $\mathbf{f}^C_n$ for each individual neighbor,
\begin{equation}
	\mathbf{f}^C_n = H_C\left(\left[ \lVert \mathbf{c} - \mathbf{c}_n \rVert, \mathbf{c} - \mathbf{c}_n, \Theta - \Theta_n, y_n, \mathbf{f}^I_n \right]^\top\right),
\end{equation}
where the angle $\Theta - \Theta_n$ is provided as direction vector and the metric features are normalized to unit length.\\

Finally, we aggregate the information from the individual neighbors into a unified representation for later processing, as shown in Figure~\ref{fig:overview}.
This requires accumulating features of an unknown (and varying) number of neighbors.
To impose no constraint on the number of features, we take inspiration from the pooling step of point-based detectors, \eg~\cite{3dod:pb:qi:pointnet}, and employ a symmetric max pooling function to form a plausibility signature $\mathbf{f}^C$ over the feature vectors of all neighbors.
Thus, in this representation the geometric relations to the surrounding objects are accumulated for the subsequent confidence estimation, taking into account the respective local properties of the neighbors.\\

\subsection{Data Fusion \& Confidence Prediction}
\label{sec:method:fusion}
To estimate the new confidence value for a detection, we merge the instance-specific and contextual geometric features.
The instance-specific information encoded in $\mathbf{f}^I$, as well as the contextual information in $\mathbf{f}^C$ are concatenated and processed using $H_F$, a MLP with sigmoid output function, to estimate the new confidence score
\begin{equation}
	\widehat{s} = H_F\left(\left[ \mathbf{f}^I, \mathbf{f}^C \right]^\top\right).
\end{equation}
We train the whole module including the two feature encoding networks $H_I$ and $H_C$ using an end-to-end training strategy.
As loss function $\mathcal{L}_{\text{conf}}(\widehat{s}, u)$ for this task we use the focal loss~\cite{2dod:loss:lin:focal-loss} to focus learning on hard examples.
The goal is to divide the set of detections as good as possible into true or false positive examples using the binary category label $u \in \{0,1\}$ as target.
As auxiliary task during training we estimate the IoU with the ground truth bounding box in a further output $\widehat{v}$ of $H_F$.
Therefore, we add an IoU-guidance~\cite{3dod:gb:yang:STD, 3dod:iou:li:gs3d, 3dod:iou:zheng:CIA-SSD} L1-loss term $\mathcal{L}_{\text{IoU}}(\widehat{v}, v)$.
This increases the importance of the point distribution statistics within the features, as they provide evidence of the bounding box accuracy from which the confidence estimate also benefits.
The overall loss function is therefore
\begin{equation}
	\mathcal{L} = \mathcal{L}_{\text{conf}}(\widehat{s}, u) \; + \; \lambda_{\text{IoU}} \; \mathcal{L}_{\text{IoU}}(\widehat{v}, v),
\end{equation}
where $\lambda_{\text{IoU}}$ is a hyper-parameter to adjust the influence of the auxiliary task.

\section{Experiments}
\label{sec:experiments}

To demonstrate the benefits of GACE, we evaluate our black-box confidence optimization method on several well-known state-of-the-art 3D object detection pipelines.
In particular, we apply it to the pillar-based PointPillars~\cite{3dod:gb:lang:PointPillars}, the voxel-based SECOND~\cite{3dod:gb:yan:SECOND} and Centerpoint~\cite{3dod:gb:yin:centerbased}, as well as the hybrid methods Part-A$^2$~\cite{3dod:hy:shi:PartA2}, PV-RCNN~\cite{3dod:hy:shi:pvrcnn} and PV-RCNN++~\cite{3dod:hy:shi:pvrcnnpp}.\\

\noindent\textbf{Datasets:} For the evaluation of our approach we use the Waymo Open Dataset~\cite{3dod:data:sun:waymo} as well as the KITTI Dataset~\cite{3dod:data:geiger:kitti}.
The Waymo dataset is one of the largest and most challenging public datasets for autonomous driving research.
It provides 798 training scenes and 202 validation scenes, where each scene consists of about 200 LiDAR samples covering the full 360$^\circ$ field-of-view.
In total, the dataset consists of over 8.9M annotated objects classified into vehicles, pedestrians and cyclists.
We follow the common evaluation protocol using the standard metrics average precision (AP), as well as average precision with heading (APH), where true positives are weighted by their heading accuracy.
In addition, we use the KITTI dataset, one of the most popular datasets for 3D object detection.
We thereby use the standard split~\cite{3dod:data:chen:kittisplit} into a training set (3712 samples) and validation set (3769 samples).
We also follow the common evaluation practice using average precision with 40 recall points.
For both datasets, the 3D IoU threshold for a true positive sample is 0.7 for vehicles and 0.5 for pedestrians and cyclists.\\

\noindent\textbf{Implementation Details: } Our experimental setup\footnote{\url{https://github.com/dschinagl/gace}} is based on the open-source toolbox OpenPCDet~\cite{other:pcdet:pcdetteam20}.
To ensure the reproducibility of our results, all base models used in the experiments have been trained on the  training set using the default configuration and default training policies of OpenPCDet.
This includes augmenting the data by randomly rotating, scaling and flipping the point cloud, as well as ground truth sampling~\cite{3dod:gb:yan:SECOND}, \ie the placement of objects from other training examples in the current frame.
To create the training data for our module, we use the base model as a black-box detection pipeline.
In a single inference run on the training set, we collect the output of the base model, \ie the set of all detections, which represents our training data.
The actual training process of our module is therefore entirely decoupled from the base model.

Our sub-networks for feature transformation, \ie~$H_I$ and the shared network $H_C$, as well as the confidence prediction network $H_F$ are two-layer MLP's with 256 feature dimensions.
The feature vector $\mathbf{f}^I$ in which the instance-specific information is encoded is 128-dimensional and $\mathbf{f}^C$ for the contextual information is 64-dimensional.
When determining the contextual geometric properties, we accumulate the neighboring objects that are within a radius of $r = 40m$.

We train our model end to end with the Adam~\cite{other:adam} optimizer and a learning rate of 0.001.
The weighting of the auxiliary loss term during training is set to $\lambda_{\text{IoU}} = 0.5$.
In all experiments we use the same architecture as well as hyperparameters, regardless of the underlying black-box base model.
We train our module over 5 epochs on a single NVIDIA\textsuperscript{\textregistered} GeForce\textsuperscript{\textregistered} RTX 3090 GPU, which takes less than 10 minutes due to our favorable lightweight model size and low feature dimensions.

\subsection{Main Results}

\begin{table*}
	\vspace{-1mm}
	\begin{center}
		\setlength\tabcolsep{3.5pt}
		\scalebox{0.8}{
			\definecolor{lightblue}{rgb}{0.748, 0.935, 0.99}
\begin{tabular}{l||cccc|cccc|cccc|cccc}
	\hline
	\multirow{3}{*}{Method} &
	\multicolumn{4}{c|}{Overall} &
	\multicolumn{4}{c|}{Vehicles (IoU=0.7)} & 
	\multicolumn{4}{c|}{Pedestrians (IoU=0.5)} & 
	\multicolumn{4}{c}{Cyclists (IoU=0.5)} \\
	& \multicolumn{2}{c}{LEVEL\_1} & \multicolumn{2}{c|}{LEVEL\_2} & 
	\multicolumn{2}{c}{LEVEL\_1} & \multicolumn{2}{c|}{LEVEL\_2} &  
	\multicolumn{2}{c}{LEVEL\_1} & \multicolumn{2}{c|}{LEVEL\_2} & 
	\multicolumn{2}{c}{LEVEL\_1} & \multicolumn{2}{c}{LEVEL\_2} \\
	& mAP & mAPH & mAP & mAPH & 
	AP & APH & AP & APH  & 
	AP & APH & AP & APH & 
	AP & APH & AP & APH \\
	\hline 
	%\cline{2-18}
	PointPillars~\cite{3dod:gb:lang:PointPillars} & 64.72& 57.07 & 58.57& 51.73 & 70.99& 70.35& 62.79& 62.20& 66.36&47.15& 58.27& 41.32& 56.81& 53.71& 54.66& 51.67\\ 
	+ GACE (Ours) & 69.25& 61.24 & 62.98& 55.73 & 71.92& 71.28& 63.63& 63.04& 72.18&51.97& 64.06& 45.96& 63.64& 60.47& 61.25& 58.20\\ 
	\cellcolor{lightblue}\textit{Improvement} & \cellcolor{lightblue}\textbf{+4.53}& \cellcolor{lightblue}\textbf{+4.17} & \cellcolor{lightblue}\textbf{+4.41}& \cellcolor{lightblue}\textbf{+4.00} & \cellcolor{lightblue}\textbf{+0.93}& \cellcolor{lightblue}\textbf{+0.93}& \cellcolor{lightblue}\textbf{+0.84}& \cellcolor{lightblue}\textbf{+0.84}& \cellcolor{lightblue}\textbf{+5.82}& \cellcolor{lightblue}\textbf{+4.82}& \cellcolor{lightblue}\textbf{+5.79}& \cellcolor{lightblue}\textbf{+4.64}& \cellcolor{lightblue}\textbf{+6.83}& \cellcolor{lightblue}\textbf{+6.76}& \cellcolor{lightblue}\textbf{+6.59}& \cellcolor{lightblue}\textbf{+6.53}\\ 
	\hline
	SECOND~\cite{3dod:gb:yan:SECOND} & 65.13& 60.81 & 59.01& 55.12 & 70.93& 70.30& 62.65& 62.07& 65.67&54.96& 57.78& 48.25& 58.78& 57.18& 56.59& 55.05\\ 
	+ GACE (Ours) & 70.17& 66.13 & 63.74& 60.06 & 71.56& 70.92& 63.22& 62.63& 71.71&61.87& 63.27& 54.37& 67.22& 65.59& 64.73& 63.16\\ 
	\cellcolor{lightblue}\textit{Improvement} & \cellcolor{lightblue}\textbf{+5.04}& \cellcolor{lightblue}\textbf{+5.32} & \cellcolor{lightblue}\textbf{+4.73}& \cellcolor{lightblue}\textbf{+4.94} & \cellcolor{lightblue}\textbf{+0.63}& \cellcolor{lightblue}\textbf{+0.62}& \cellcolor{lightblue}\textbf{+0.57}& \cellcolor{lightblue}\textbf{+0.56}& \cellcolor{lightblue}\textbf{+6.04}& \cellcolor{lightblue}\textbf{+6.91}& \cellcolor{lightblue}\textbf{+5.49}& \cellcolor{lightblue}\textbf{+6.12}& \cellcolor{lightblue}\textbf{+8.44}& \cellcolor{lightblue}\textbf{+8.41}& \cellcolor{lightblue}\textbf{+8.14}& \cellcolor{lightblue}\textbf{+8.11}\\ 
	\hline
	Part-A$^2$~\cite{3dod:hy:shi:PartA2} & 70.30& 66.66 & 63.53& 60.27 & 73.35& 72.81& 64.73& 64.24& 70.02&61.01& 60.83& 52.85& 67.53& 66.18& 65.03& 63.73\\ 
	+ GACE (Ours) & 73.07& 69.21 & 66.24& 62.77 & 73.99& 73.43& 65.38& 64.87& 72.36&62.93& 63.21& 54.81& 72.84& 71.28& 70.13& 68.63\\ 
	\cellcolor{lightblue}\textit{Improvement} & \cellcolor{lightblue}\textbf{+2.77}& \cellcolor{lightblue}\textbf{+2.55} & \cellcolor{lightblue}\textbf{+2.71}& \cellcolor{lightblue}\textbf{+2.50} & \cellcolor{lightblue}\textbf{+0.64}& \cellcolor{lightblue}\textbf{+0.62}& \cellcolor{lightblue}\textbf{+0.65}& \cellcolor{lightblue}\textbf{+0.63}& \cellcolor{lightblue}\textbf{+2.34}& \cellcolor{lightblue}\textbf{+1.92}& \cellcolor{lightblue}\textbf{+2.38}& \cellcolor{lightblue}\textbf{+1.96}& \cellcolor{lightblue}\textbf{+5.31}& \cellcolor{lightblue}\textbf{+5.10}& \cellcolor{lightblue}\textbf{+5.10}& \cellcolor{lightblue}\textbf{+4.90}\\ 
	\hline
	Centerpoint~\cite{3dod:gb:yin:centerbased} & 73.01& 70.35 & 66.79& 64.33 & 72.87& 72.33& 64.76& 64.27& 74.48&68.22& 66.55& 60.81& 71.69& 70.50& 69.06& 67.92\\ 
	+ GACE (Ours) & 75.58& 72.98 & 69.19& 66.76 & 74.49& 73.99& 66.19& 65.73& 78.62&72.48& 70.44& 64.73& 73.64& 72.48& 70.94& 69.83\\ 
	\cellcolor{lightblue}\textit{Improvement} & \cellcolor{lightblue}\textbf{+2.57}& \cellcolor{lightblue}\textbf{+2.63} & \cellcolor{lightblue}\textbf{+2.40}& \cellcolor{lightblue}\textbf{+2.43} & \cellcolor{lightblue}\textbf{+1.62}& \cellcolor{lightblue}\textbf{+1.66}& \cellcolor{lightblue}\textbf{+1.43}& \cellcolor{lightblue}\textbf{+1.46}& \cellcolor{lightblue}\textbf{+4.14}& \cellcolor{lightblue}\textbf{+4.26}& \cellcolor{lightblue}\textbf{+3.89}& \cellcolor{lightblue}\textbf{+3.92}& \cellcolor{lightblue}\textbf{+1.95}& \cellcolor{lightblue}\textbf{+1.98}& \cellcolor{lightblue}\textbf{+1.88}& \cellcolor{lightblue}\textbf{+1.91}\\ 
	PV-RCNN~\cite{3dod:hy:shi:pvrcnn} & 71.11& 66.78 & 64.40& 60.50 & 74.79& 74.17& 66.17& 65.61& 72.06&61.46& 63.00& 53.56& 66.48& 64.72& 64.03& 62.34\\ 
	+ GACE (Ours) & 73.38& 68.89 & 66.65& 62.57 & 75.20& 74.55& 66.57& 65.97& 73.84&62.97& 64.88& 55.14& 71.12& 69.16& 68.50& 66.60\\ 
	\cellcolor{lightblue}\textit{Improvement} & \cellcolor{lightblue}\textbf{+2.27}& \cellcolor{lightblue}\textbf{+2.11} & \cellcolor{lightblue}\textbf{+2.25}& \cellcolor{lightblue}\textbf{+2.07} & \cellcolor{lightblue}\textbf{+0.41}& \cellcolor{lightblue}\textbf{+0.38}& \cellcolor{lightblue}\textbf{+0.40}& \cellcolor{lightblue}\textbf{+0.36}& \cellcolor{lightblue}\textbf{+1.78}& \cellcolor{lightblue}\textbf{+1.51}& \cellcolor{lightblue}\textbf{+1.88}& \cellcolor{lightblue}\textbf{+1.58}& \cellcolor{lightblue}\textbf{+4.64}& \cellcolor{lightblue}\textbf{+4.44}& \cellcolor{lightblue}\textbf{+4.47}& \cellcolor{lightblue}\textbf{+4.26}\\
	PV-RCNN++~\cite{3dod:hy:shi:pvrcnnpp} & 75.72& 73.05 & 69.22& 66.73 & 77.30& 76.81& 68.92& 68.47& 78.91&72.42& 70.43& 64.41& 70.95& 69.90& 68.31& 67.31\\ 
	+ GACE (Ours) & 76.76& 74.02 & 70.31& 67.75 & 77.42& 76.93& 69.05& 68.59& 79.59&72.97& 71.34& 65.18& 73.26& 72.16& 70.54& 69.48\\ 
	\cellcolor{lightblue}\textit{Improvement} & \cellcolor{lightblue}\textbf{+1.04}& \cellcolor{lightblue}\textbf{+0.97} & \cellcolor{lightblue}\textbf{+1.09}& \cellcolor{lightblue}\textbf{+1.02} & \cellcolor{lightblue}\textbf{+0.12}& \cellcolor{lightblue}\textbf{+0.12}& \cellcolor{lightblue}\textbf{+0.13}& \cellcolor{lightblue}\textbf{+0.12}& \cellcolor{lightblue}\textbf{+0.68}& \cellcolor{lightblue}\textbf{+0.55}& \cellcolor{lightblue}\textbf{+0.91}& \cellcolor{lightblue}\textbf{+0.77}& \cellcolor{lightblue}\textbf{+2.31}& \cellcolor{lightblue}\textbf{+2.26}& \cellcolor{lightblue}\textbf{+2.23}& \cellcolor{lightblue}\textbf{+2.17}\\ 
	\hline
\end{tabular}
		}
	\end{center}
	\caption{Performance comparison on the Waymo Open Dataset~\cite{3dod:data:sun:waymo} validation set for different baseline methods with and without our confidence enhancement module GACE.}
	\label{tab:waymo3d}
\end{table*}

Table~\ref{tab:waymo3d} summarizes the results of our confidence optimization on the Waymo dataset for different baseline detection pipelines.
For each detector architecture, we report the performance of the base detector as well as the results after applying our GACE module.
Please note that for better traceability, the results of the base detectors correspond to the respective OpenPCDet implementations (see OpenPCDet modelzoo).
Some reported baseline scores are even higher than in their original papers due to augmentation techniques used in OpenPCDet.

For all baseline detectors and all classes, adjusting the confidence level with our approach leads to an increase in performance without exception.
The overall performance gains range from +1.02mAPH (LEVEL\_2) for PV-RCNN++ to +4.94mAPH (LEVEL\_2) for SECOND, demonstrating the significance of this geometric information in estimating a confidence score for a detection.\\

\noindent
\textbf{Object Classes: }
The performance improvements are significantly higher for the classes of pedestrians and cyclists.
Intuitively, objects of these vulnerable road users contain significantly fewer 3D points due to their smaller spatial extent compared to vehicles.
This effect is even aggravated by possible occlusions.
Furthermore, the class of cyclists in particular is underrepresented in the training data.
These properties make objects of these classes more difficult to detect, which leads to a higher number of false positives, but also makes confidence estimation more complex.
Especially in these cases, our confidence enhancement method benefits from the additional geometric information when separating false positives from true positives, resulting in a higher impact of GACE.

\noindent
\textbf{Base Detectors: }
The overall performance gain is highest when applying GACE on pillar-based (PointPillars~\cite{3dod:gb:lang:PointPillars}) and voxel-based methods (SECOND~\cite{3dod:gb:yan:SECOND} and Centerpoint~\cite{3dod:gb:yin:centerbased}). 
These methods lose valuable point information already in their preprocessing stage, \ie~by voxelization.
Thus, the optimization potential for these methods is higher than for the hybrid methods, which explicitly incorporate point-level information at detection locations. 
However, by explicitly exploiting both instance-specific and contextual geometric properties of the detections, GACE also leads to a significant performance improvement for these methods, \ie~Part-A$^2$~\cite{3dod:hy:shi:PartA2}, PV-RCNN~\cite{3dod:hy:shi:pvrcnn} and PV-RCNN++~\cite{3dod:hy:shi:pvrcnnpp}, most notably for the classes of pedestrians and cyclists.

\begin{figure*}[t]
	\includegraphics[width=\linewidth]{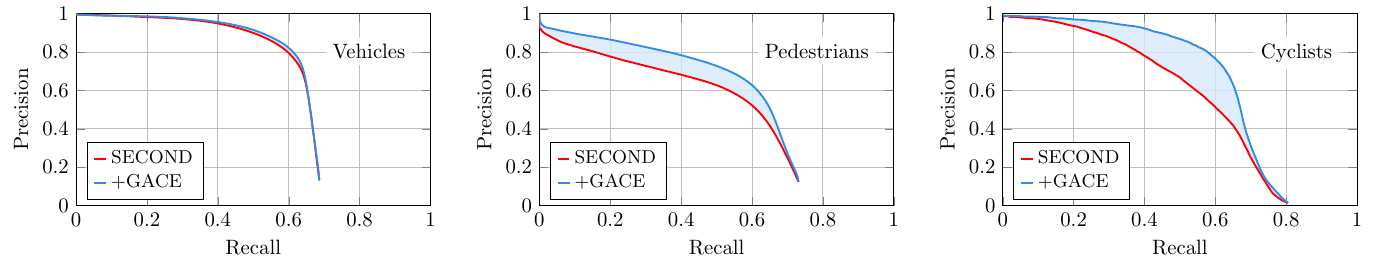}
	\caption{
		Precision-recall plots for a SECOND~\cite{3dod:gb:yan:SECOND} model as base detector with and without our confidence enhancement module on the Waymo Open Dataset~\cite{3dod:data:sun:waymo} validation set for LEVEL\_2 APH.
		Especially at higher recall levels, GACE's better separation of true and false positives leads to higher precision.
	}
	\label{fig:precision_recall}
\end{figure*}

\noindent
\textbf{Precision/Recall Plots: }
To better illustrate the impact of our method, Figure~\ref{fig:precision_recall} shows the precision/recall plots before and after applying GACE with a SECOND~\cite{3dod:gb:yan:SECOND} model as the base detector (See the supplemental material for the other base detectors).
Since recall remains unaffected, the significant performance gains of GACE are entirely due to an increase in precision.
The better separation of true and false positives is mainly seen in the more challenging classes of pedestrians and cyclists, where the number of false detections is significantly higher and confidence estimation is more difficult.
Furthermore, we see that the precision gains are higher in regions of higher recall, \ie for detections with a lower confidence score.
Especially these initially underrated detections of objects of the vulnerable classes represent a safety risk that can be reduced by GACE.

\noindent
\textbf{Range-based Evaluation: }
Table~\ref{tab:waymo3drange} summarizes our evaluation across the different distance ranges of Waymo.
This also shows that GACE can consistently improve the performance of all baseline detectors and in all sub-ranges, especially in the far range.
Detections at long distances are more challenging due to the lower point density, making it harder to distinguish true positives from false positives.
Therefore, the geometric information is even more valuable in these cases, \eg from +2.93mAPH (LEVEL\_2) for PV-RCNN++ up to +4.97mAPH (LEVEL\_2) for PointPillars.
The detailed evaluation results for all subclasses can be found in the supplemental material.

\begin{table}
	\vspace{-1mm}
	\begin{center}
		\setlength\tabcolsep{3.5pt}
		\scalebox{0.8}{
			\definecolor{lightblue}{rgb}{0.748, 0.935, 0.99}
\begin{tabular}{l||ccc}
	\hline
	\multirow{3}{*}{Method} & \multicolumn{3}{c}{Overall (LEVEL\_2)}\\
	& 0-30m & 30-50m & 50m-Inf\\
	& mAP / mAPH & mAP / mAPH &	mAP / mAPH\\
	\hline 
	%\cline{2-18}
	PointPillars~\cite{3dod:gb:lang:PointPillars} & 75.10 / 67.59  & 55.31 / 48.53  & 37.18 / 31.00  \\
	+ GACE (Ours) & 78.32 / 70.68  & 60.17 / 53.01  & 42.88 / 35.97  \\
	\cellcolor{lightblue}\textit{Improvement} & \cellcolor{lightblue}\textbf{+3.22} / \textbf{+3.09}  & \cellcolor{lightblue}\textbf{+4.86} / \textbf{+4.48}  & \cellcolor{lightblue}\textbf{+5.70} / \textbf{+4.97}  \\ 
	\hline
	SECOND~\cite{3dod:gb:yan:SECOND} & 75.55 / 71.87  & 56.55 / 52.42  & 37.73 / 33.39  \\
	+ GACE (Ours) & 78.64 / 75.39  & 60.10 / 56.13  & 41.69 / 37.39  \\
	\cellcolor{lightblue}\textit{Improvement} & \cellcolor{lightblue}\textbf{+3.09} / \textbf{+3.52}  & \cellcolor{lightblue}\textbf{+3.55} / \textbf{+3.71}  & \cellcolor{lightblue}\textbf{+3.96} / \textbf{+4.00}  \\ 
	\hline
	Part-A$^2$~\cite{3dod:hy:shi:PartA2} & 79.93 / 76.93  & 61.77 / 58.01  & 40.13 / 36.44  \\
	+ GACE (Ours) & 81.62 / 78.52  & 64.63 / 60.59  & 45.06 / 40.94  \\
	\cellcolor{lightblue}\textit{Improvement} & \cellcolor{lightblue}\textbf{+1.69} / \textbf{+1.59}  & \cellcolor{lightblue}\textbf{+2.86} / \textbf{+2.58}  & \cellcolor{lightblue}\textbf{+4.93} / \textbf{+4.50}  \\ 
	\hline
	Centerpoint~\cite{3dod:gb:yin:centerbased} & 81.29 / 79.12  & 65.45 / 62.66  & 46.45 / 43.27  \\
	+ GACE (Ours) & 82.94 / 80.90  & 67.56 / 64.76  & 50.20 / 46.80  \\
	\cellcolor{lightblue}\textit{Improvement} & \cellcolor{lightblue}\textbf{+1.65} / \textbf{+1.78}  & \cellcolor{lightblue}\textbf{+2.11} / \textbf{+2.10}  & \cellcolor{lightblue}\textbf{+3.75} / \textbf{+3.53}  \\ 
	\hline
	PV-RCNN~\cite{3dod:hy:shi:pvrcnn} & 79.77 / 76.24  & 62.57 / 58.10  & 42.51 / 37.77  \\
	+ GACE (Ours) & 81.25 / 77.66  & 64.76 / 60.11  & 47.01 / 41.90  \\
	\cellcolor{lightblue}\textit{Improvement} & \cellcolor{lightblue}\textbf{+1.48} / \textbf{+1.42}  & \cellcolor{lightblue}\textbf{+2.19} / \textbf{+2.01}  & \cellcolor{lightblue}\textbf{+4.50} / \textbf{+4.13}  \\ 
	\hline
	PV-RCNN++~\cite{3dod:hy:shi:pvrcnnpp} & 83.08 / 80.98  & 67.86 / 65.04  & 48.69 / 45.26  \\
	+ GACE (Ours) & 83.59 / 81.43  & 68.69 / 65.83  & 51.81 / 48.19  \\
	\cellcolor{lightblue}\textit{Improvement} & \cellcolor{lightblue}\textbf{+0.51} / \textbf{+0.45}  & \cellcolor{lightblue}\textbf{+0.83} / \textbf{+0.79}  & \cellcolor{lightblue}\textbf{+3.12} / \textbf{+2.93}  \\ 
	\hline
\end{tabular}
		}
	\end{center}
	\caption{Consistent performance gains on the Waymo Open Dataset~\cite{3dod:data:sun:waymo} validation set across distance ranges for different baseline methods for LEVEL\_2 over all classes.}
	\label{tab:waymo3drange}
\end{table}

\subsection{Ablation Studies}
\noindent \textbf{Main Components:} We evaluate the contribution of our proposed instance-specific and contextual properties as well as the influence of the auxilary IoU loss for a SECOND~\cite{3dod:gb:yan:SECOND} model as baseline in Table~\ref{tab:ablation}.
Incorporating the instance-specific geometric properties already leads to a significant performance increase of +3.65AP/+3.32APH.
Including also the contextual information, \ie incorporating the relationships to the neighboring detection hypotheses further increases the performance by another +0.71AP/+1.14APH.
Compared to the contributions of these two components, there is only a minor impact of incorporating the IoU guidance, leading to additional +0.37AP/+0.48APH.
Note that the degradation when adding $\mathcal{L}_{\text{IoU}}$ to only contextual features is caused by the lack of point information, which does not allow a reasonable estimation of the IoU.
Overall, the instance-specific geometric information contributes stronger than the contextual geometric information.\\

\begin{table}
	\centering
	\def\arraystretch{1.2}
    \setlength\tabcolsep{3.5pt}
	\resizebox{\linewidth}{!}{
		\begin{tabular}{ccccccc}
			\multirow{2}{*}{Baseline} & Instance & \multirow{2}{*}{Contextual} & Aux-Loss  & LEVEL\_2 & LEVEL\_2   \\
			& Specific & & $\mathcal{L}_{\text{IoU}}$ & mAP & mAPH \\
			\hline
			$\checkmark$ & & & & 59.01 & 55.12   \\
			$\checkmark$ & $\checkmark$ & & & 62.66 & 58.44  \\
            $\checkmark$ & & $\checkmark$ & &  60.33 & 56.90 \\
			$\checkmark$ & $\checkmark$ & $\checkmark$ & & 63.37 & 59.58 \\
            $\checkmark$ & $\checkmark$ &  &$\checkmark$ & 62.94 & 58.78 \\
            $\checkmark$ & & $\checkmark$  &$\checkmark$ & 60.19 & 56.77 \\
			$\checkmark$ & $\checkmark$ & $\checkmark$ &$\checkmark$ & 63.74 & 60.06 \\
		\end{tabular}
	}
	\vspace{2mm}
	\caption{		
		Ablation experiments for a SECOND~\cite{3dod:gb:yan:SECOND} model as base detector over all classes on the Waymo Open Dataset~\cite{3dod:data:sun:waymo} validation set.
		We show the impact of each GACE component.\\
	}
	\label{tab:ablation}
\end{table}

\noindent \textbf{Instance-Specific Properties:} 
We analyze the contribution of each feature group within the instance-specific properties, namely \textit{box properties}~$(\mathbf{b},\lVert \mathbf{c} \rVert)$, \textit{number of points}~$(|\mathcal{X}_{\mathbf{b}}|)$, \textit{viewing angle}~$(\alpha)$, and \textit{point statistics}~$(\mathcal{X}_{\mathbf{b}}^{\text{mean}},  \mathcal{X}_{\mathbf{b}}^{\text{std}}, \mathcal{X}_{\mathbf{b}}^{\text{min}}, \mathcal{X}_{\mathbf{b}}^{\text{max}})$.
Table~\ref{tab:abl_inst_prop} shows the impact of each group on the overall performance when used exclusively, indicating a high contribution of box properties and point statistics.
A complete list of the combinations can be found in the supplemental material.\\

\begin{table}[t]
    \centering
    \setlength\tabcolsep{3.5pt}
    \scalebox{0.8}{
        \begin{tabular}{p{2.5cm}|p{1.975cm}|p{2.5275cm}|p{2.1cm}}
            \multirowcell{1}{{Box Properties}}  & \multirowcell{1}{\# Points} & \multirowcell{1}{{Viewing Angle}} & \multirowcell{1}{{Point Stat.}} \\
            \hline
            \multirowcell{1}{$56.98^\textbf{\small \color{OliveGreen}+1.86}$} & \multirowcell{1}{$55.96^\textbf{\small \color{OliveGreen}+0.84}$} & \multirowcell{1}{$55.86^\textbf{\small \color{OliveGreen}+0.74}$} & \multirowcell{1}{$58.15^\textbf{\small \color{OliveGreen}+3.03}$} \\
        \end{tabular}}
    \vspace{2mm}
    \caption{Impact of the different instance-specific feature groups when used exclusively for a SECOND model as base detector on the Waymo Open Dataset~\cite{3dod:data:sun:waymo} over all classes (LEVEL\_2 mAPH).}
    \label{tab:abl_inst_prop}
\end{table}

\begin{figure}[t]
	\centering
    \includegraphics[width=0.98\linewidth]{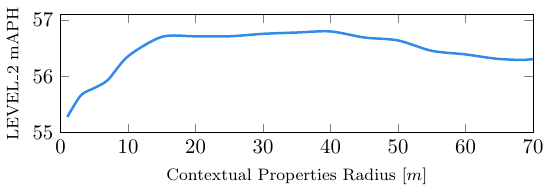}
    \caption{Performance of purely context-based GACE for different radius $r$ (SECOND base model / Waymo overall).}
	\label{fig:radius}
\end{figure}

\noindent \textbf{Contextual Properties Radius:}
The dependence of the performance on the chosen context radius $r$ is shown in Figure~\ref{fig:radius}.
It can be seen that the performance increases significantly up to $\sim15m$, followed by a slight degradation starting at $\sim40m$, which illustrates the importance of neighboring objects in the near and middle ranges. \\

\subsection{Generalization and Transferability}

Since we use only the detection attributes and corresponding properties of the underlying point cloud as input, a GACE module trained on base detector A can be directly applied to another base detector B.
Furthermore, since all metric features are normalized to their maximum possible value and the statistical parameters (point distribution) are computed from a normalized unit-length box, GACE can be applied not only to a different detector, but also directly to a different detector on a different dataset.
Therefore, to demonstrate the general applicability of GACE, we freeze a GACE module trained on detections from a SECOND detector on Waymo, and apply it directly to detections of a SECOND detector on the KITTI dataset~\cite{3dod:data:geiger:kitti}.
As shown in Table~\ref{tab:kitti3d}, first row, this also leads to considerable performance improvements despite the distribution shift to a different dataset (different LiDAR sensor, different country).
Even more remarkably, significant performance gains are also obtained when the same GACE module is applied even to a different base detector, \eg +7.42AP for PV-RCNN on pedestrians while achieving the same performance for cars.
The results demonstrate the excellent generalization capability of GACE and the general validity of the geometric information regardless of the dataset.

\begin{table}[t]
	\vspace{-1mm}
	\centering
	\setlength\tabcolsep{3.5pt}
	\scalebox{0.8}{
		\begin{tabular}{l|c|c|c}
			Method & Car & Pedestrian & Cyclist \\ \hline
			SECOND (KITTI) & $81.61$ & $51.14$ & $66.74$ \\ 
			+GACE (SECOND Waymo) & $82.04^\textbf{\small \color{OliveGreen}+0.43}$ & $57.11^\textbf{\small \color{OliveGreen}+5.97}$ & $70.99^\textbf{\small \color{OliveGreen}+4.25}$ \\ 
			\hline
			PointPillars (KITTI) & $78.39$ & $51.41$ & $62.94$ \\ 
			+GACE (SECOND Waymo) & $78.51^\textbf{\small \color{OliveGreen}+0.12}$ & $55.30^\textbf{\small \color{OliveGreen}+3.89}$ & $67.94^\textbf{\small \color{OliveGreen}+5.00}$ \\
			\hline
			Part-A$^2$ (KITTI) & $82.92$ & $59.73$ & $70.10$ \\ 
			+GACE (SECOND Waymo) & $82.94^\textbf{\small \color{OliveGreen}+0.02}$ & $64.21^\textbf{\small \color{OliveGreen}+4.48}$ & $72.16^\textbf{\small \color{OliveGreen}+2.06}$ \\ 
			\hline
			PV-RCNN (KITTI) & $82.86$ & $53.64$ & $70.42$ \\
			+GACE (SECOND Waymo) & $82.84^\textbf{\small \color{Red}-0.02}$ & $61.06^\textbf{\small \color{OliveGreen}+7.42}$ & $72.70^\textbf{\small \color{OliveGreen}+2.28}$ \\
		\end{tabular}
	}
	\vspace{2mm}
	\caption{Model Transfer: Applying a GACE module trained on SECOND detections on Waymo (without LiDAR elongation, which is not available on KITTI) to different detectors on the KITTI dataset (moderate difficulty / @R40).}
	\label{tab:kitti3d}
\end{table}

\subsection{Runtime Analysis}
In Table~\ref{tab:runtime} we show the runtime analysis of GACE for a 360$^\circ$ field-of-view Waymo point cloud using a single NVIDIA\textsuperscript{\textregistered} GeForce\textsuperscript{\textregistered} RTX 3090 GPU. 
The computationally most intensive part is the feature extraction, especially extracting the statistical features for all detections.
However, this can be solved efficiently via PyTorch \texttt{einsum}.
During inference, we first compute the instance-specific plausibility $\mathbf{f}^I$ for each detection via $H_I$, which is then used as input to $H_C$ for the corresponding neighbors.
Overall, GACE is capable of processing $\sim490$ Waymo point clouds per second with $\sim100$ detections each.

\begin{table}[t]
    \vspace{-1mm}
    \centering
    \scalebox{0.8}{
        \begin{tabular}{p{2cm}|p{5cm}|r}
            \multirowcell{3}[0pt][l]{\textbf{Feature} \\ \textbf{Extraction}} & Points in Boxes Query & $0.56$ ms \\
            & Geometric \& Statistical Features & $0.98$ ms \\
            & Neighboring Object ID's Query & $0.07$ ms \\
            \hline
            \multirowcell{3}[0pt][l]{\textbf{Network} \\ \textbf{Inference}} & $H_I$ (Instance-specific) & 0.14 ms\\
            & $H_C$ (Contextual) & 0.17 ms\\
            & $H_F$ (Confidence Estimation) & 0.12 ms \\
            \hline
            \multicolumn{1}{l}{\textbf{Overall}} & \multicolumn{2}{r}{\textbf{{2.04 ms}}} \\ 
        \end{tabular}}
    \vspace{2mm}
    \caption{GACE runtime analysis per 360$^\circ$ field-of-view Waymo point cloud on a single NVIDIA\textsuperscript{\textregistered} GeForce\textsuperscript{\textregistered} RTX 3090 GPU.}
    \label{tab:runtime}
\end{table}

\section{Limitations}

While our method has proven effective in improving object detection performance for vulnerable classes such as pedestrians and cyclists, it is worth noting that it may not offer such significant benefits for simpler classes such as vehicles.
This is because all detectors typically have few false positives for these \textit{easier-to-detect} objects.
Therefore, there is less room for improvement for this class.
However, it is important to consider the context in which our method is applied.
Even if the performance improvement is not as large for simpler classes, the overall impact of our method on safety should not be underestimated.
In many real-world scenarios, pedestrians and cyclists are particularly vulnerable, and any improvement in their detection can make a significant contribution to safety.

\section{Conclusion}

We proposed GACE, a method to better evaluate object hypotheses from black-box LiDAR-based 3D object detectors by explicitly assessing numerous geometric properties inherent in the detections.
This enables a better separation between true and false positive detections and thus significantly improves the performance.
In a comprehensive analysis, we were able to show the performance of our method for several state-of-the-art detectors and also demonstrate the generalisation capabilities of GACE. This underlines the importance and general validity of the geometric information inherent in 3D detections, which has been largely neglected in the past.\\

{
\small
\noindent\textbf{Acknowledgements}
This work was partially funded by the Austrian FFG project iLIDS4SAM~(878713) and by the Christian Doppler Laboratory for Embedded Machine Learning.
}

{\small
\bibliographystyle{ieee_fullname}
\bibliography{gace.bib}
}
\clearpage
\newpage
\appendix
\section{Supplemental Material}

This supplemental material provides additional insights and quantitative results, as well as potential limitations of our confidence enhancement approach.\\

\subsection{Geometric Dependencies}
\begin{figure}[th!]
	\centering
	\includegraphics[width=0.8\linewidth]{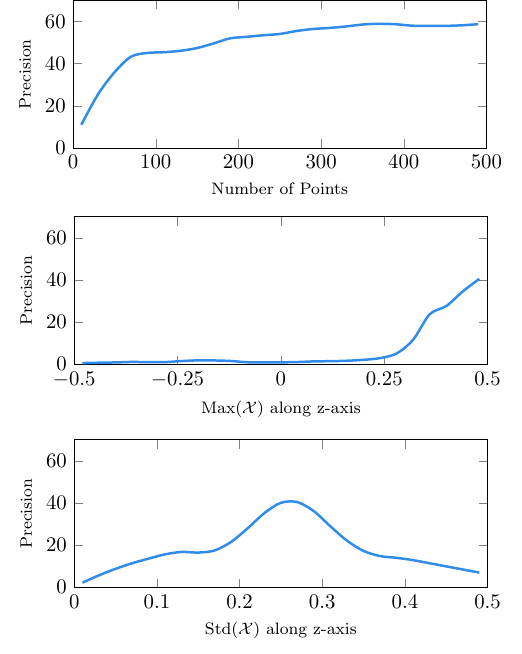}
    \caption{Precision of a SECOND~\cite{3dod:gb:yan:SECOND} model on the Waymo Open Datase~\cite{3dod:data:sun:waymo} for the pedestrian class on different low-level point statistics. Note that the statistics refer to zero-centered boxes on all axes, normalized to unit length.}
	\label{fig:geom_dep}
\end{figure}

Figure~\ref{fig:geom_dep} shows an additional example of how the detection performance depends on geometric properties.
Here, we show how well a vanilla SECOND~\cite{3dod:gb:yan:SECOND} model detects pedestrians on the Waymo dataset \wrt low-level statistics of the points within the detection bounding boxes.
Besides the expected influence of the number of points (top), we found that the detection accuracy also strongly depends on the distribution of points along the $z\text{-axis}$ of the bounding box,~\ie~along the height of the suspected pedestrian.
On the one hand, it can be seen that pedestrians are only detected reliably if points are present in the upper quarter of the bounding box,~\ie~in the area of the head-shoulder silhouette (middle).
On the other hand, the precision increases the better the points are distributed across the entire $z\text{-axis}$ (bottom).
The incorporation of these simple properties therefore enables an improved assessment of the confidence score.

\subsection{Waymo Test Set Results}
To demonstrate the reliability of GACE, we also evaluate our approach on the Waymo Open Dataset~\cite{3dod:data:sun:waymo} test set using the official evaluation server.
Note that for Waymo, detectors typically perform better on the test set compared to the larger validation set.
Despite the already improved performance of the base detectors, GACE enables comparable and consistent improvements across the different detectors and classes, see Table~\ref{tab:waymo3dtestset}.

\subsection{Ablation Study Instance-Specific Properties}
In Table~\ref{tab:ablation_is}, we show the complete list of all combinations of instance-specific properties and the impact on the overall performance.
We analyze the contribution of each feature group within the instance-specific properties, namely \textit{box properties}~$(\mathbf{b},\lVert \mathbf{c} \rVert)$, \textit{number of points}~$(|\mathcal{X}_{\mathbf{b}}|)$, \textit{viewing angle}~$(\alpha)$, and \textit{point statistics}~$(\mathcal{X}_{\mathbf{b}}^{\text{mean}},  \mathcal{X}_{\mathbf{b}}^{\text{std}}, \mathcal{X}_{\mathbf{b}}^{\text{min}}, \mathcal{X}_{\mathbf{b}}^{\text{max}})$.

\begin{table}
    \centering
    \def\arraystretch{1.2}
    \setlength\tabcolsep{3.5pt}
    \scalebox{0.89}{
    \resizebox{\linewidth}{!}{
        \begin{tabular}{ccccccc}
            Box & \multirow{2}{*}{\# Points} & Viewing & Point & LEVEL\_2 &  \multirow{2}{*}{Improvement}   \\
            Properties & & Angle & statistics & mAPH &\\
            \hline
            $\checkmark$ & $\checkmark$ & $\checkmark$ & $\checkmark$ & 58.78 & $\textbf{\color{OliveGreen}+3.66}$\\
            $\checkmark$ & $\checkmark$ & $\checkmark$ & & 58.00 & $\textbf{\color{OliveGreen}+2.88}$\\
            $\checkmark$ & $\checkmark$ & & $\checkmark$ & 58.55 & $\textbf{\color{OliveGreen}+3.43}$\\
            $\checkmark$ & $\checkmark$ & & & 57.33 & $\textbf{\color{OliveGreen}+2.21}$\\
            $\checkmark$ & & $\checkmark$ & $\checkmark$ & 58.46 & $\textbf{\color{OliveGreen}+3.34}$\\
            $\checkmark$ & & $\checkmark$ & & 57.21 & $\textbf{\color{OliveGreen}+2.09}$\\
            $\checkmark$ & & & $\checkmark$ & 58.49 & $\textbf{\color{OliveGreen}+3.37}$\\
            $\checkmark$ & & & & 56.98 & $\textbf{\color{OliveGreen}+1.86}$\\
            & $\checkmark$ & $\checkmark$ & $\checkmark$ & 58.26 & $\textbf{\color{OliveGreen}+3.14}$\\
            & $\checkmark$ & $\checkmark$ & & 56.04 & $\textbf{\color{OliveGreen}+0.92}$\\
            & $\checkmark$ & & $\checkmark$ & 58.23 & $\textbf{\color{OliveGreen}+3.11}$\\
            & $\checkmark$ & & & 55.96 & $\textbf{\color{OliveGreen}+0.84}$\\
            & & $\checkmark$ & $\checkmark$ & 58.30 & $\textbf{\color{OliveGreen}+3.18}$\\
            & & $\checkmark$ & & 55.86 & $\textbf{\color{OliveGreen}+0.74}$\\
            & & & $\checkmark$ & 58.15 & $\textbf{\color{OliveGreen}+3.03}$\\
        \end{tabular}
        }
    }
    \vspace{2mm}
    \caption{Impact of instance-specific feature groups (SECOND base model / Waymo LEVEL\_2 mAPH overall).}	
    \label{tab:ablation_is}
\end{table}

\begin{table*}[t]
	\begin{center}
		\setlength\tabcolsep{3.5pt}
		\scalebox{0.72}{
			\definecolor{lightblue}{rgb}{0.748, 0.935, 0.99}
\begin{tabular}{l||ccc|ccc|ccc}
	\hline
	\multirow{3}{*}{Method} & \multicolumn{3}{c|}{Vehicle (LEVEL\_2)}& \multicolumn{3}{c|}{Pedestrian (LEVEL\_2)}& \multicolumn{3}{c}{Cyclist (LEVEL\_2)}\\
	& 0-30m & 30-50m & 50m-Inf& 0-30m & 30-50m & 50m-Inf& 0-30m & 30-50m & 50m-Inf\\
	& mAP / mAPH & mAP / mAPH &	mAP / mAPH & mAP / mAPH & mAP / mAPH &	mAP / mAPH& mAP / mAPH & mAP / mAPH &	mAP / mAPH\\
	\hline 
	%\cline{2-18}
	PointPillars~\cite{3dod:gb:lang:PointPillars} & 88.10 / 87.58  & 61.88 / 61.18  & 34.78 / 34.01  & 67.80 / 49.03  & 58.32 / 40.98  & 41.77 / 27.60  & 69.40 / 66.17  & 45.71 / 43.43  & 35.00 / 31.39 \\
	+ GACE (Ours) & 88.46 / 87.96  & 63.03 / 62.31  & 36.12 / 35.31  & 72.37 / 53.01  & 64.15 / 45.81  & 49.12 / 33.25  & 74.12 / 71.08  & 53.34 / 50.92  & 43.40 / 39.35 \\
	\cellcolor{lightblue}\textit{Improvement} & \cellcolor{lightblue}\textbf{+0.36} / \textbf{+0.38}  & \cellcolor{lightblue}\textbf{+1.15} / \textbf{+1.13}  & \cellcolor{lightblue}\textbf{+1.34} / \textbf{+1.30}  & \cellcolor{lightblue}\textbf{+4.57} / \textbf{+3.98}  & \cellcolor{lightblue}\textbf{+5.83} / \textbf{+4.83}  & \cellcolor{lightblue}\textbf{+7.35} / \textbf{+5.65}  & \cellcolor{lightblue}\textbf{+4.72} / \textbf{+4.91}  & \cellcolor{lightblue}\textbf{+7.63} / \textbf{+7.49}  & \cellcolor{lightblue}\textbf{+8.40} / \textbf{+7.96} \\ 
	\hline
	SECOND~\cite{3dod:gb:yan:SECOND} & 88.50 / 87.98  & 62.14 / 61.49  & 33.93 / 33.20  & 66.64 / 57.51  & 58.20 / 47.88  & 41.36 / 31.23  & 71.53 / 70.14  & 49.31 / 47.90  & 37.88 / 35.74 \\
	+ GACE (Ours) & 88.81 / 88.28  & 62.98 / 62.32  & 35.01 / 34.23  & 71.23 / 63.27  & 62.65 / 52.81  & 45.64 / 35.58  & 75.90 / 74.61  & 54.65 / 53.26  & 44.41 / 42.35 \\
	\cellcolor{lightblue}\textit{Improvement} & \cellcolor{lightblue}\textbf{+0.31} / \textbf{+0.30}  & \cellcolor{lightblue}\textbf{+0.84} / \textbf{+0.83}  & \cellcolor{lightblue}\textbf{+1.08} / \textbf{+1.03}  & \cellcolor{lightblue}\textbf{+4.59} / \textbf{+5.76}  & \cellcolor{lightblue}\textbf{+4.45} / \textbf{+4.93}  & \cellcolor{lightblue}\textbf{+4.28} / \textbf{+4.35}  & \cellcolor{lightblue}\textbf{+4.37} / \textbf{+4.47}  & \cellcolor{lightblue}\textbf{+5.34} / \textbf{+5.36}  & \cellcolor{lightblue}\textbf{+6.53} / \textbf{+6.61} \\ 
	\hline
	Part-A$^2$~\cite{3dod:hy:shi:PartA2} & 89.53 / 89.09  & 64.90 / 64.34  & 37.11 / 36.42  & 71.03 / 63.65  & 61.27 / 52.00  & 41.92 / 33.33  & 79.24 / 78.04  & 59.13 / 57.68  & 41.37 / 39.58 \\
	+ GACE (Ours) & 89.87 / 89.43  & 65.72 / 65.12  & 38.10 / 37.30  & 72.50 / 64.91  & 63.65 / 53.90  & 46.25 / 36.71  & 82.49 / 81.22  & 64.52 / 62.75  & 50.82 / 48.80 \\
	\cellcolor{lightblue}\textit{Improvement} & \cellcolor{lightblue}\textbf{+0.34} / \textbf{+0.34}  & \cellcolor{lightblue}\textbf{+0.82} / \textbf{+0.78}  & \cellcolor{lightblue}\textbf{+0.99} / \textbf{+0.88}  & \cellcolor{lightblue}\textbf{+1.47} / \textbf{+1.26}  & \cellcolor{lightblue}\textbf{+2.38} / \textbf{+1.90}  & \cellcolor{lightblue}\textbf{+4.33} / \textbf{+3.38}  & \cellcolor{lightblue}\textbf{+3.25} / \textbf{+3.18}  & \cellcolor{lightblue}\textbf{+5.39} / \textbf{+5.07}  & \cellcolor{lightblue}\textbf{+9.45} / \textbf{+9.22} \\ 
	\hline
	Centerpoint~\cite{3dod:gb:yin:centerbased} & 88.88 / 88.39  & 65.26 / 64.72  & 37.05 / 36.50  & 74.44 / 69.48  & 66.95 / 60.42  & 51.89 / 44.21  & 80.56 / 79.50  & 64.15 / 62.85  & 50.40 / 49.09 \\
	+ GACE (Ours) & 89.88 / 89.43  & 66.73 / 66.20  & 38.61 / 38.05  & 77.80 / 73.17  & 70.86 / 64.22  & 56.33 / 48.14  & 81.13 / 80.10  & 65.09 / 63.86  & 55.66 / 54.22 \\
	\cellcolor{lightblue}\textit{Improvement} & \cellcolor{lightblue}\textbf{+1.00} / \textbf{+1.04}  & \cellcolor{lightblue}\textbf{+1.47} / \textbf{+1.48}  & \cellcolor{lightblue}\textbf{+1.56} / \textbf{+1.55}  & \cellcolor{lightblue}\textbf{+3.36} / \textbf{+3.69}  & \cellcolor{lightblue}\textbf{+3.91} / \textbf{+3.80}  & \cellcolor{lightblue}\textbf{+4.44} / \textbf{+3.93}  & \cellcolor{lightblue}\textbf{+0.57} / \textbf{+0.60}  & \cellcolor{lightblue}\textbf{+0.94} / \textbf{+1.01}  & \cellcolor{lightblue}\textbf{+5.26} / \textbf{+5.13} \\ 
	\hline
	PV-RCNN~\cite{3dod:hy:shi:pvrcnn} & 89.87 / 89.40  & 66.39 / 65.74  & 39.73 / 38.88  & 71.84 / 63.21  & 63.47 / 52.72  & 46.60 / 35.87  & 77.58 / 76.12  & 57.86 / 55.85  & 41.20 / 38.56 \\
	+ GACE (Ours) & 90.04 / 89.56  & 67.07 / 66.35  & 40.42 / 39.48  & 73.11 / 64.35  & 65.16 / 54.13  & 49.87 / 38.31  & 80.59 / 79.07  & 62.06 / 59.84  & 50.74 / 47.91 \\
	\cellcolor{lightblue}\textit{Improvement} & \cellcolor{lightblue}\textbf{+0.17} / \textbf{+0.16}  & \cellcolor{lightblue}\textbf{+0.68} / \textbf{+0.61}  & \cellcolor{lightblue}\textbf{+0.69} / \textbf{+0.60}  & \cellcolor{lightblue}\textbf{+1.27} / \textbf{+1.14}  & \cellcolor{lightblue}\textbf{+1.69} / \textbf{+1.41}  & \cellcolor{lightblue}\textbf{+3.27} / \textbf{+2.44}  & \cellcolor{lightblue}\textbf{+3.01} / \textbf{+2.95}  & \cellcolor{lightblue}\textbf{+4.20} / \textbf{+3.99}  & \cellcolor{lightblue}\textbf{+9.54} / \textbf{+9.35} \\ 
	\hline
	PV-RCNN++~\cite{3dod:hy:shi:pvrcnnpp} & 90.99 / 90.58  & 69.74 / 69.24  & 43.18 / 42.56  & 78.71 / 73.77  & 70.48 / 63.60  & 54.87 / 46.64  & 79.54 / 78.58  & 63.35 / 62.28  & 48.02 / 46.57 \\
	+ GACE (Ours) & 91.15 / 90.74  & 69.86 / 69.35  & 43.36 / 42.69  & 79.22 / 74.14  & 71.34 / 64.31  & 56.75 / 48.10  & 80.39 / 79.39  & 64.88 / 63.83  & 55.33 / 53.77 \\
	\cellcolor{lightblue}\textit{Improvement} & \cellcolor{lightblue}\textbf{+0.16} / \textbf{+0.16}  & \cellcolor{lightblue}\textbf{+0.12} / \textbf{+0.11}  & \cellcolor{lightblue}\textbf{+0.18} / \textbf{+0.13}  & \cellcolor{lightblue}\textbf{+0.51} / \textbf{+0.37}  & \cellcolor{lightblue}\textbf{+0.86} / \textbf{+0.71}  & \cellcolor{lightblue}\textbf{+1.88} / \textbf{+1.46}  & \cellcolor{lightblue}\textbf{+0.85} / \textbf{+0.81}  & \cellcolor{lightblue}\textbf{+1.53} / \textbf{+1.55}  & \cellcolor{lightblue}\textbf{+7.31} / \textbf{+7.20} \\ 
	\hline
\end{tabular}
		}
	\end{center}
	\caption{Performance comparison on the Waymo Open Dataset~\cite{3dod:data:sun:waymo} validation set across distance ranges for different baseline methods with and without our confidence enhancement module for difficulty LEVEL\_2 and for all classes.}
	\label{tab:waymo3drange_detail}
\end{table*}

\subsection{Range-based Evaluation}
Table~\ref{tab:waymo3drange_detail} presents the detailed results of the range-based evaluation on the Waymo Open Dataset~\cite{3dod:data:sun:waymo} for all individual classes.
Adjusting the confidence values using GACE leads to an improvement in performance for all classes and across all distance ranges.
The detailed results for each class also show that our confidence enhancement approach achieves the highest performance gain at long ranges.
These more stable detections are particularly important for safety, as they give autonomous vehicles more time to react, avoid collisions and plan a safe and efficient route.
Nevertheless, as can be seen in the example of the pedestrian class for SECOND~\cite{3dod:gb:yan:SECOND} and Centerpoint~\cite{3dod:gb:yin:centerbased}, performance can also be significantly improved for close-range detections. 

\begin{table}[t]
	\centering
	\setlength\tabcolsep{3pt}
	\scalebox{0.73}{
		\begin{tabular}{l|c|c|c|c}
		\multirow{2}{*}{Method} &
		\multicolumn{2}{c|}{LEVEL\_1} & \multicolumn{2}{c}{LEVEL\_2} \\
		& mAP & mAPH & mAP & mAPH  \\ \hline
		PART-A$^2$ & $71.66$ & $67.99$ & $65.84$ & $62.47$ \\
		+GACE & $73.42^\textbf{\small \color{OliveGreen}+1.76}$ & $69.65^\textbf{\small \color{OliveGreen}+1.66}$ & $67.55^\textbf{\small \color{OliveGreen}+1.71}$ & $64.08^\textbf{\small \color{OliveGreen}+1.61}$ \\ \hline
		
		Centerpoint & $74.39$ & $71.67$ & $68.92$ & $66.38$ \\
		+GACE & $76.03^\textbf{\small \color{OliveGreen}+1.64}$ & $73.39^\textbf{\small \color{OliveGreen}+1.72}$ & $70.45^\textbf{\small \color{OliveGreen}+1.53}$ & $67.97^\textbf{\small \color{OliveGreen}+1.59}$ \\ \hline
		
		PV-RCNN & $73.21$ & $68.84$ & $67.42$ & $63.39$ \\
		+GACE  & $74.61^\textbf{\small \color{OliveGreen}+1.40}$ & $70.14^\textbf{\small \color{OliveGreen}+1.30}$ & $68.79^\textbf{\small \color{OliveGreen}+1.37}$ & $64.65^\textbf{\small \color{OliveGreen}+1.26}$ \\		
		\end{tabular}
	}
	\vspace{2mm}
	\caption{Performance gains using GACE on the Waymo Open Dataset~\cite{3dod:data:sun:waymo} test set over all classes.}
	\label{tab:waymo3dtestset}
\end{table}

\subsection{Geometric Dependencies with GACE}
\begin{figure}[t]
	\centering
    \includegraphics[width=0.84\linewidth]{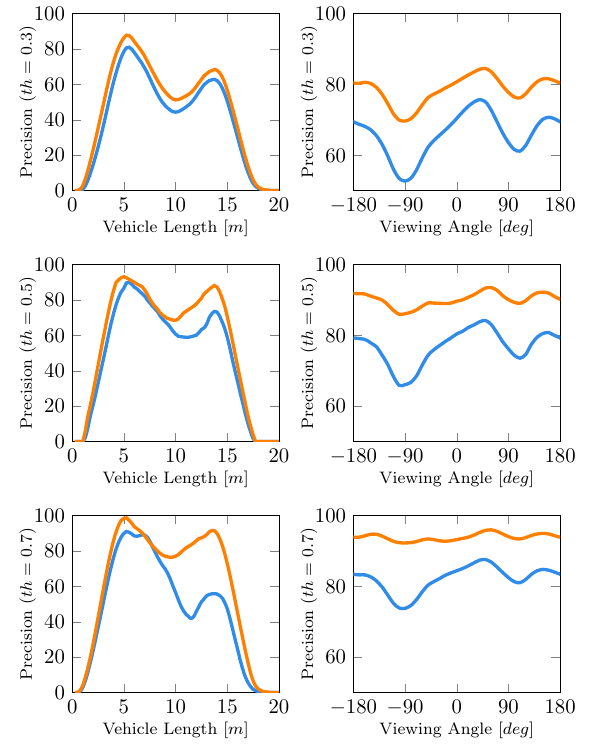}
    \vspace{-1mm}
    \caption{Precision of a SECOND model on the Waymo Open Dataset for the vehicle class as a function of the object length (left column) and of the viewing angle (right column), before (blue) and after (orange) applying GACE for different confidence score thresholds (rows).}
	\label{fig:geometry_gace_before_after}
\end{figure}

In Figure~\ref{fig:geometry_gace_before_after} we show the impacts of GACE on the geometric dependencies of the precision of a SECOND model on the Waymo dataset using the vehicle length (left column) and the viewing angle (right column) as exmaples.
Therefore, we thresholded the detections at different confidence thresholds of 0.3, 0.5, and 0.7 (rows) and show the precision before (blue) and after (orange) applying our confidence enhancement.
GACE can especially improve the precision for challenging samples (such as trucks, \ie~long vehicles), which are typically underrepresented (and thus, suffer from low confidence scores).
Thus, this is only partially reflected by the evaluation metrics (\eg~improvement for \textit{vehicles} with SECOND is +0.56 LEVEL\_2 APH).

\subsection{Precision-Recall Plots}
\begin{figure*}
    \begin{center}
        \includegraphics[width=0.85\linewidth]{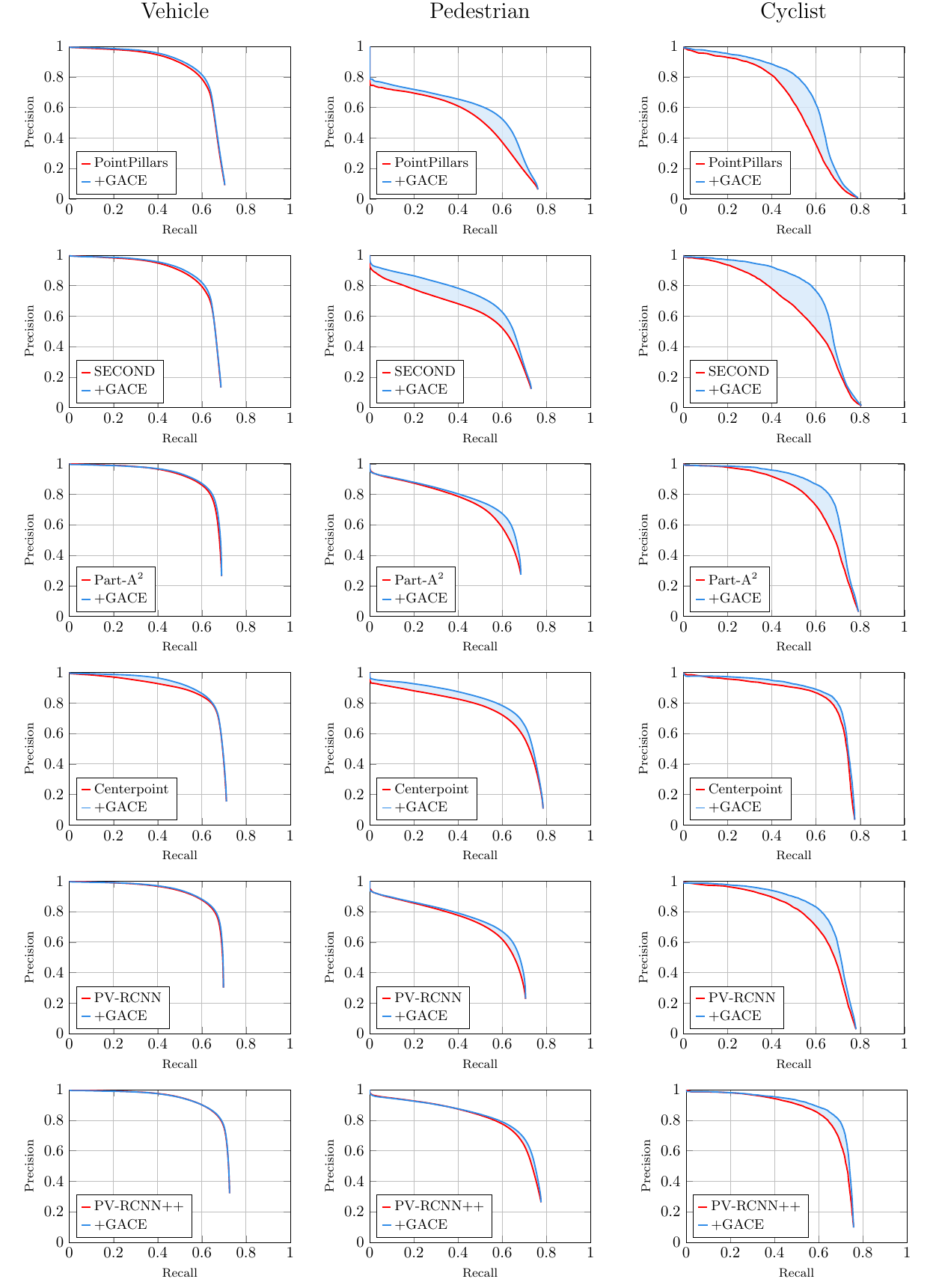}
    \end{center}
	\vspace{-0.25cm}
	\caption{Precision-recall plots for all base models and all classes on the Waymo Open Dataset~\cite{3dod:data:sun:waymo} validation set for LEVEL\_2 APH.
		GACE~allows us to better separate true and false positive hypotheses, which leads to more stable detection results, especially for the vulnerable road users.}
	\label{fig:pr_plots}
\end{figure*}

Figure~\ref{fig:pr_plots} shows the remaining precision-recall plots of the evaluation on the Waymo validation set for each class at LEVEL\_2 APH difficulty level.
The plots confirm the performance improvements especially for the spatially smaller and therefore more difficult classes of pedestrians and cyclists. 
In addition, it is noticeable that the increase in precision is particularly high in the regions with a high recall, \ie for detections with a low confidence score.

\subsection{Qualitative Results}

\begin{figure*}[!t]
	\centering
    \includegraphics[width=0.75\linewidth]{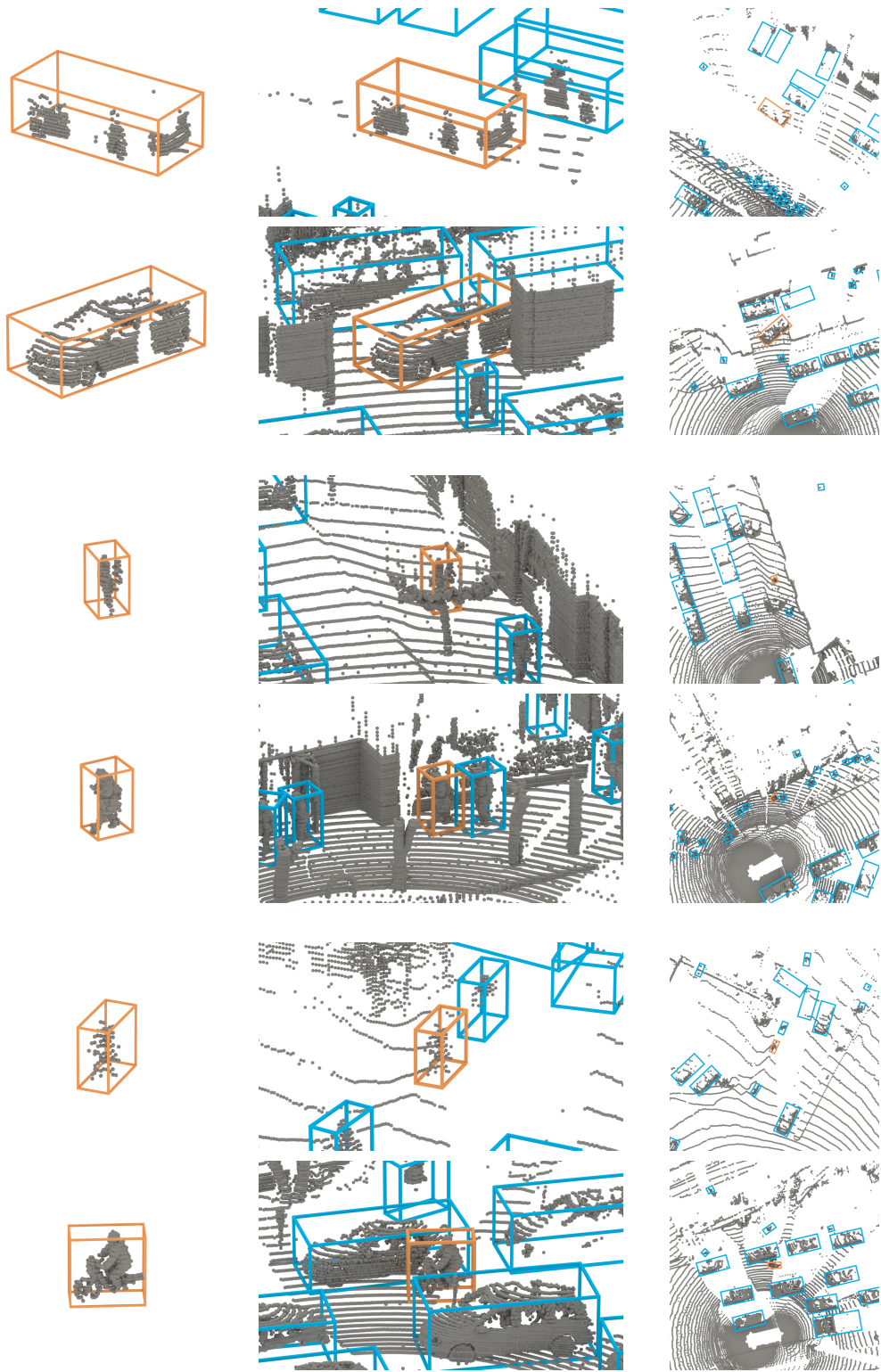}
    \vspace{2mm}
    \caption{Qualitative Examples (Highest Score Increase): For each Waymo class, we show the two instances in the validation set where GACE increased the confidence value the most compared to the baseline SECOND model (vehicle at the top, pedestrian in the middle and cyclist at the bottom rows). Best viewed on screen.}
	\label{fig:increase}
\end{figure*}

\begin{figure*}[!t]
	\centering
    \includegraphics[width=0.75\linewidth]{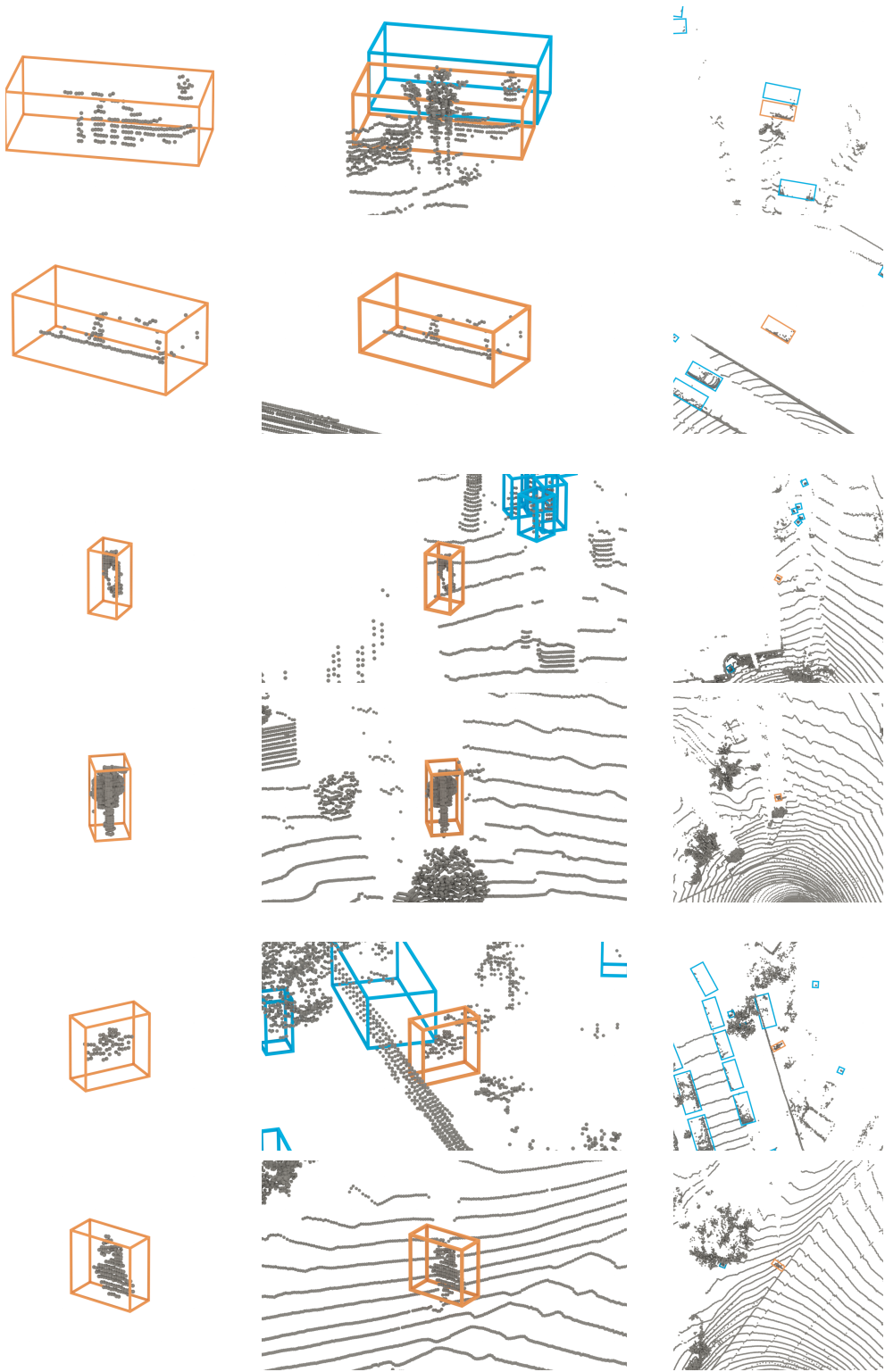}
    \vspace{2mm}
    \caption{Qualitative Examples (Highest Score Decrease): For each Waymo class, we show the two instances in the validation set where GACE decreased the confidence value the most compared to the baseline SECOND model (vehicle at the top, pedestrian in the middle and cyclist at the bottom rows). Best viewed on screen.}
	\label{fig:decrease}
\end{figure*}

To illustrate the behavior of GACE using qualitative examples, Figure~\ref{fig:increase} shows the two instances of each Waymo class in the validation set where GACE increased the confidence value the most compared to the baseline SECOND model. Conversely, Figure~\ref{fig:decrease} shows the two instances per class where the confidence score decreased the most.

\end{document}